\documentclass[review]{elsarticle}

\usepackage{hyperref}

\journal{Journal of \LaTeX\ Templates}

\newcommand{\etal}{\textit{et al}. }
\newcommand{\ie}{\textit{i}.\textit{e}., }
\newcommand{\eg}{\textit{e}.\textit{g}. }
\usepackage{mathtools}
\usepackage{algorithm}
\usepackage{algorithmic}
\usepackage{multirow}
\usepackage{array}
\usepackage{color}

\begin{document}

\begin{frontmatter}

\title{Deep-Person: Learning Discriminative Deep Features for Person Re-Identification}

%% Group authors per affiliation:
\author{Xiang Bai,
Mingkun Yang,
Tengteng Huang,\\
Zhiyong Dou,
Rui Yu,
Yongchao Xu\corref{cor}}
\address{School of Electronic Information and Communications, Huazhong University
of Science and Technology (HUST), Wuhan, 430074, China}

%% or include affiliations in footnotes:
\cortext[cor]{Corresponding author}
\ead{\{xbai,yangmingkun,tengtenghuang,zydou\}@hust.edu.cn,yurui.thu@gmail.com, yongchaoxu@hust.edu.cn}

\begin{abstract}

Person re-identification (Re-ID) requires discriminative features focusing on the full person to cope with inaccurate person bounding box detection, background clutter, and occlusion. Many recent person Re-ID methods attempt to learn such features describing full person details via part-based feature representation. However, the spatial context between these parts is ignored for the independent extractor on each separate part. In this paper, we propose to apply Long Short-Term Memory (LSTM) in an end-to-end way to model the pedestrian, seen as a sequence of body parts from head to foot. Integrating the contextual information strengthens the discriminative ability of local feature aligning better to full person. We also leverage the complementary information between local and global feature. Furthermore, we integrate both identification task and ranking task in one network, where a discriminative embedding and a similarity measurement are learned concurrently. This results in a novel three-branch framework named Deep-Person, which learns highly discriminative features for person Re-ID. Experimental results demonstrate that Deep-Person outperforms the state-of-the-art methods by a large margin on three challenging datasets including Market-1501, CUHK03, and DukeMTMC-reID. Code is available at: \href{https://github.com/zydou/Deep-Person}{https://github.com/zydou/Deep-Person}.
%Specifically, combining with a re-ranking approach, we achieve a \textbf{90.84\%} mAP on Market-1501 under single query setting.
\end{abstract}


\end{frontmatter}

% \linenumbers

\section{Introduction} \label{sec:introduction}

Person re-identification (Re-ID) refers the task of matching a specific person across multiple non-overlapping cameras. It has been receiving increasing attention in the computer vision community thanks to its various surveillance applications. Despite decades of study on person Re-ID task, it is still very challenging due to inaccurate person bounding box detection and large variations in illumination, pose, background clutter, occlusion, and ambiguity in visual appearance. Discriminative features focusing mainly on full person are inevitable to cope with these challenges in person Re-ID.

Most early works in person Re-ID either focus on discriminative hand-craft feature representation or robust distance metric for similarity measurement. Benefiting from the development of deep learning and increasing large-scale datasets~\cite{zheng2017unlabeled,zheng2015scalable,li2014deepreid}, recent person Re-ID methods combine feature extraction and distance metric into an end-to-end deep convolution neural network (CNN). Nevertheless, most recent CNN-based methods endeavor to either design a better feature representation or develop a more robust feature learning, but rarely both aspects together. {Recently, some semi-supervised and unsupervised methods are proposed to further promote this field~\cite{WuLDYBY19,FanZYY18}, which achieve satisfactory performance with few or even no labels.}

\begin{figure}[!t]
\begin{center}
   \includegraphics[width=\linewidth,height=0.4\textheight]{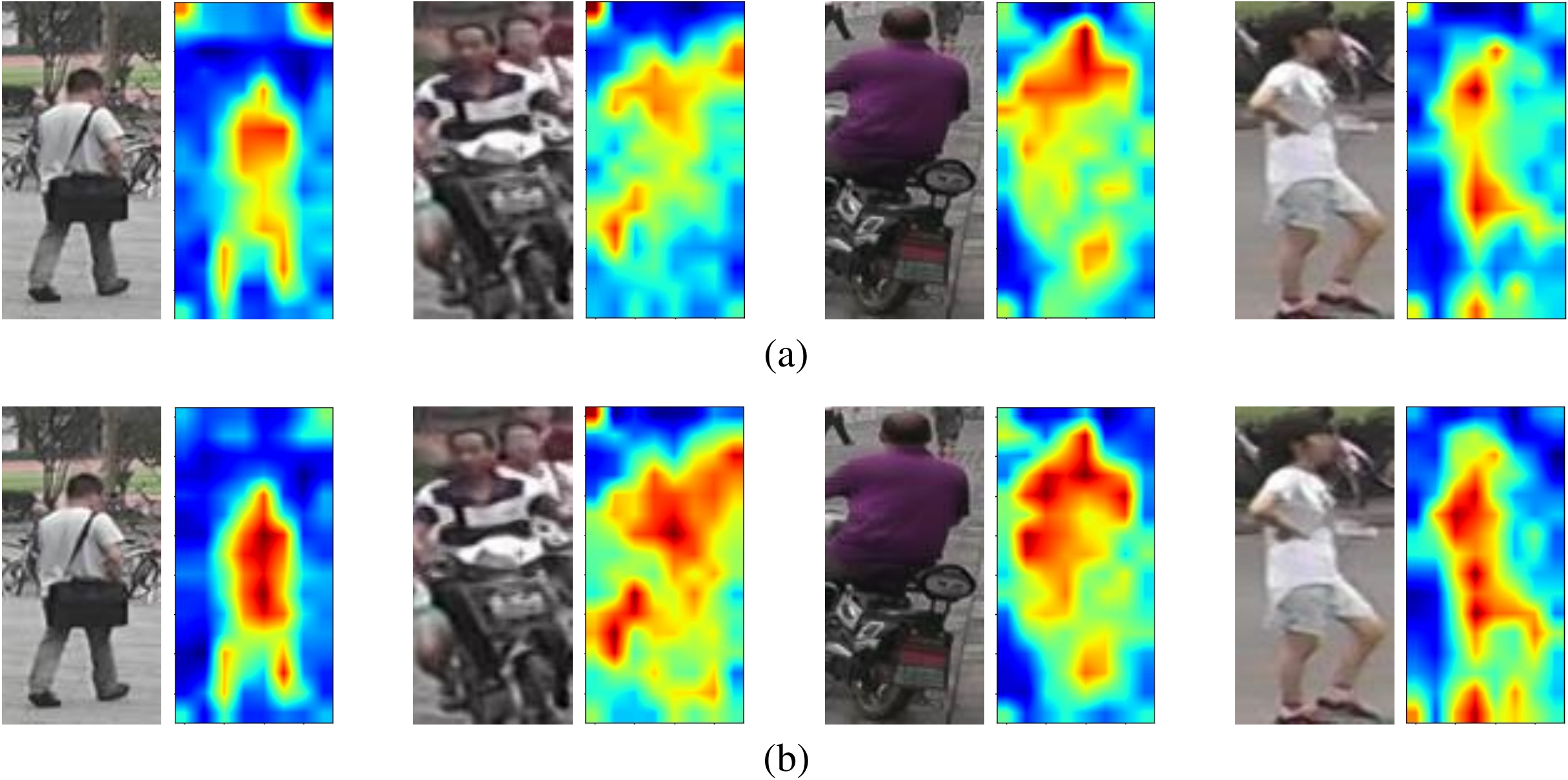}
\end{center}
  \caption{Visualization of CNN learned feature maps with traditional global-based method (a) and our proposed Deep-Person (b). The heat map reveals regions that the network focuses on to compute feature for person Re-ID. Deep-Person tends to mainly focus on the whole body with accurate and detailed features aligned to person. Conversely, without the context between each part, the traditional one suffers from occlusion, blurring, and background clutter.}
\label{fig:misalignment}
\end{figure}

The CNN-based methods focusing on better feature representations can be roughly divided into three categories: 1) Global full-body representation, which is adopted in many methods~\cite{li2014deepreid,chen2017multi}. Global average pooling is widely used for such global feature extraction, which decreases the granularity of features, thus resulting in missing local details (see Fig.~\ref{fig:misalignment} (a)); 2) Local body-part representation, which has been exploited in many works with variant part partitions. A straightforward partition into predefined rigid body parts is used in many works~\cite{li2017JLML,ustinova2017multi,ahmed2015improved,BenGZJWM19}. This may make the learned feature focus on some person details, Yet, due to pose variations, imperfect pedestrian detectors, and occlusion, such trivial partition fails to correctly learn features aligned to full person, leading to part-based features far from robust.
Some recent works endeavor to develop better body partitions with some sophisticated methods~\cite{yao2017deep,li2017learning,wu2018what} or using extra pose annotation~\cite{zhao2017spindle,su2017pose}. Although these part-based methods can enrich the generated feature describing better some person details, they all ignore the contextual information between the body parts, still failing to well align to full person and suffering from occlusion, blurring, and background noise. In~\cite{varior2016siamese}, the authors propose to first convert the original person image into sequential LOMO and Color Names features, then rely on Recurrent Neural Network (RNN) to model the spatial context. Yet, the separation of sequence feature extraction and spatial context modeling hinders the end-to-end training and optimization, resulting in degraded performance; 3) Combination of global and local representation~\cite{li2017learning,li2017JLML,wu2016enhanced, yan2018multi}, which concatenates the global and part-based feature as the final feature representation. This combined feature representation usually requires more computation and extra space in test phase due to an extra branch compared to the single branch model, yielding slower runtime in practice.

The methods dedicated for robust feature learning usually consider the person Re-ID problem as either a classification task or a ranking task. Thanks to recently increasing large-scale Re-ID datasets, person Re-ID is regarded as a multi-class person identification task in many works~\cite{li2014deepreid,zheng2016wild,zheng2016person}. The obtained ID-discriminative Embedding (IDE) given by the penultimate fully connected layer has shown great potentials for person Re-ID. Yet, the training objective of identification task is not totally consistent with the testing procedure. Such learned IDE may be optimal for identification model on the training set, but may not be optimal to describe unseen person images in the test stage. Furthermore, the identification task does not explicitly learn a similarity measurement required for retrieval during the test stage. On the other hand, the ranking task aims to make the distance between the positive pairs closer than that of the negative pairs by a given margin. Therefore, a similarity measurement is explicitly learned. Yet, all identity information of the annotated Re-ID is not fully utilized. Recently, some works~\cite{chen2017multi,wang2016joint,liu2017end} leverage both classification and ranking tasks with triplet loss to learn more discriminative features for person Re-ID. 

\begin{figure*}
\begin{center}
\includegraphics[width=\linewidth,,height=0.35\textheight]{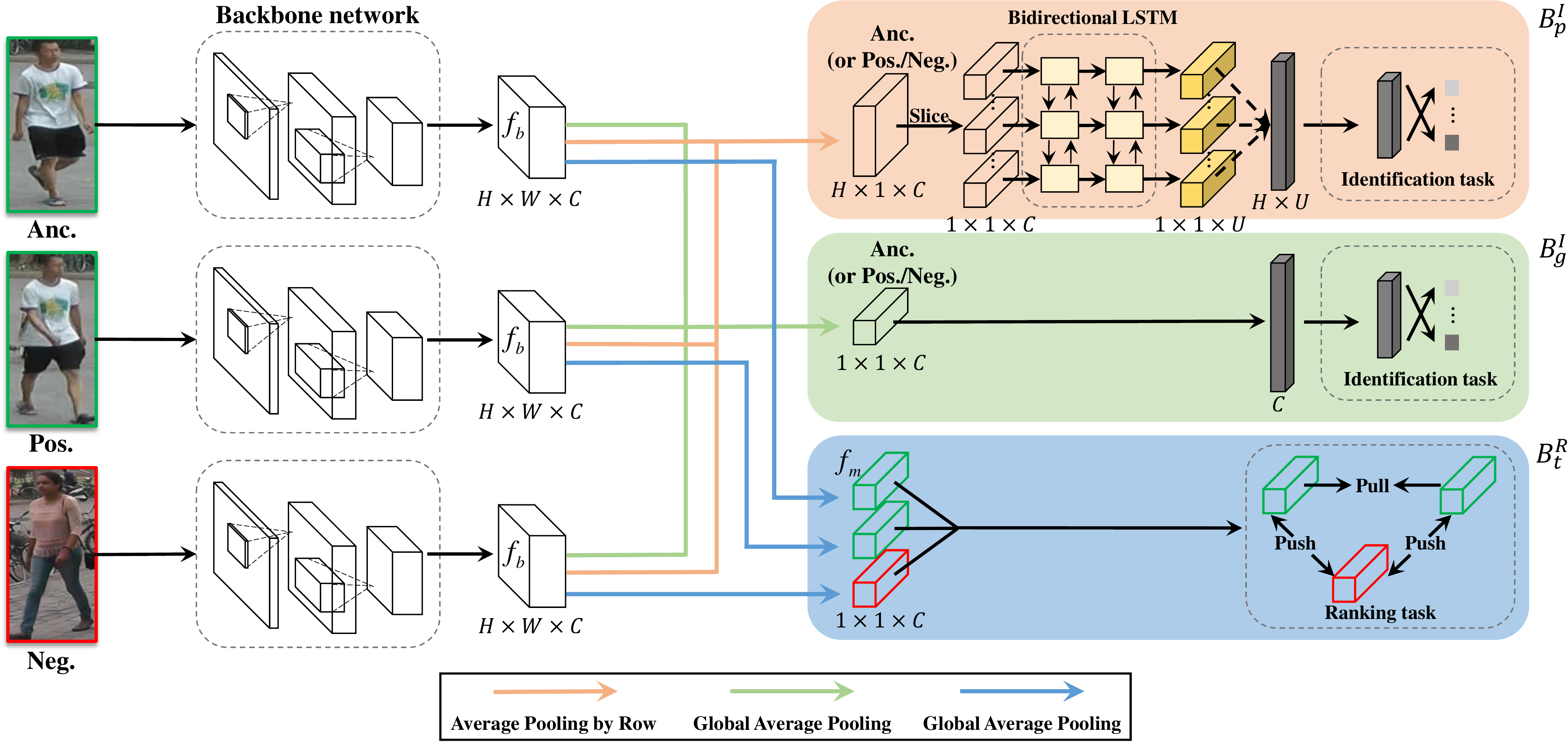}
\end{center}
   \caption{Illustration of the Deep-Person architecture. Given a triplet of images $T$, each image is fed into a part-based identification branch $B_p^I$ and a global-based identification branch $B_g^I$. Meanwhile, a distance ranking branch $B_t^R$ is applied on $T$ using the triplet loss function. Note that each image in $T$ is fed into the same backbone network. The duplication of the backbone is only for visualization purpose. The two colors of Global Average Pooling are the same. Different colors are only used to represent different branches.}
\label{fig:pipeline}
\end{figure*}

In this paper, we propose to model the pedestrian as a sequence of body parts from head to foot, thus each part is strongly related to other ones. However, the existing part-based methods usually ignore this contextual information. To capture the context within the body sequence, we employ Long Short-Term Memory (LSTM)~\cite{hochreiter1997long} in an end-to-end way, enhancing the discriminative capacity of local feature which aligns better to full person with the prior knowledge of body structure.
% In this paper, we propose to model the pedestrian as a sequence of body parts from head to foot, and learn all part features together with the spatial contextual information rather than use an independent branch for each separate part. For that, we apply Long Short-Term Memory (LSTM)~\cite{hochreiter1997long} in an end-to-end way, enhancing the discriminative capacity of local feature which aligns better to full person with the prior knowledge of body structure.
We also adopt global full-body representation in the proposed Deep-Person model. We feed the global and part-based features into two separate network branches for identification tasks. Different from the classical combination of global and local representation~\cite{cheng2016person,li2017learning,li2017JLML} concatenating global and part-based feature as the final feature representation for person Re-ID, we further add a ranking task branch using triplet loss to explicitly learn the similarity measurement. More specifically, we use the global average pooled feature $f_m$ of the backbone feature $f_b$ (shown in Fig.~\ref{fig:pipeline}) as similarity estimation, also considered as the final pedestrian descriptor. Such three-branch Deep-Person model learns highly discriminative features for person Re-ID. 

% The main contributions of this paper are three folds: 1) We propose to regard the pedestrian as a sequence of body parts from head to foot, and apply LSTM in an end-to-end fashion to take into account the contextual information between body parts, enhancing the discriminative capacity of local feature which aligns better to full person; 2) We develop a novel three-branch framework which leverages two kinds of complementary advantages: local body-part and global full-body feature representation, as well as identification task and ranking task for a better feature learning. The proposed Deep-Person yields highly discriminative features for person Re-ID; 3) In the test phase, the proposed Deep-Person only performs a forward pass of the backbone network followed by a global average pooling. Consequently, compared to the single branch model, our model requires no additional runtime and space during testing, while still outperforming the state-of-the-art methods by large margins on three popular Re-ID datasets.

The main contributions of this paper are three folds: 1) We propose to regard the pedestrian as a sequence of body parts from head to foot, and apply LSTM to take into account the contextual information between body parts, while most existing part-based methods ignore the contexts. Such contextual information helps to learn more discriminative features focusing on the human body; 2) We develop a novel three-branch framework which leverages two kinds of complementary advantages: local body-part and global full-body feature representation, as well as identification task and ranking task for a better feature learning. Existing works make use of either complementary information, rarely both of them to obtain more discriminative features for Person Re-ID. 3) In the test phase, the proposed Deep-Person only performs a forward pass of the backbone network followed by a global average pooling. Consequently, compared to the single branch model, our model requires no additional runtime and space during testing, while still outperforming the state-of-the-art methods by large margins on three popular Re-ID datasets.

\section{Related Work}

Some methods consider the person Re-ID as a special case of image retrieval problem, \ie given a probe image, the framework ranks all gallery images based on their distances in the projected space with the probe, then returns the top $k$ most similar images. In this sense, they tend to focus on robust distance metrics, such as~\cite{wang2018equidistance,zhao2017multiple,liu2018m3l,ZHAO201879,CHENG2018,ZHAO201890}. Yet in this paper, we concentrate on a high quality pedestrian descriptor through better feature representation and more robust feature learning. So we focus on two types of closely related deep learning methods for person Re-ID: those relying on part-based representations and those focusing on multi-loss learning. For a complete review of person Re-ID methods, the interested reader is referred to~\cite{zheng2016person,KaranamGWRCR19}. One novelty of the proposed Deep-Person lies on the use of LSTM to model the pedestrian seen as a sequence of body parts from head to foot. We also shortly review some related work using LSTM for sequence modeling.
% In this section, we focus on the deep learning based methods which are most relevant to our proposed algorithm.

\paragraph{Part-based person Re-ID approaches}
There are many methods use part-based representation to learn discriminative features for person Re-ID. Following the strategy of part partition, the part-based approaches can be roughly divided into two categories: 1) Rigid body part partition, which has been widely adopted in many methods~\cite{cheng2016person,chen2016similarity,varior2016siamese,fendri2018multi}, where the authors use predefined rigid grids as local parts. Each part is fed into an individual branch. All the individual part features are then concatenated together as the final part-based representation. For example, PCB~\cite{pcb} vertically splits the final feature maps into several stripes, each of which represents certain body part of the input person image on the corresponding receptive field. Moreover, attention mechanism is applied on each stripe to make the part features focus on the body parts. The relationship within different body parts is ignored in PCB. 2) Flexible body part partition, which is more reasonable to localize appropriate parts. For example, Yao~\etal\cite{yao2017deep} use an unsupervised method to generate a set of part boxes, and then employ the RoI pooling to produce part features. The Spatial Transformer Networks (STN)~\cite{jaderberg2015spatial} with novel spatial constraints are applied to localize deformable person parts in~\cite{li2017learning}. With extra pose annotation, Zhao~\etal\cite{zhao2017spindle} utilize the learned body joints to obtain subregion bounding boxes. Su~\etal\cite{su2017pose} further extend~\cite{zhao2017spindle} by normalizing the pose-based parts into fixed size and orientation, and introducing the Pose Transformation Network (PTN) to eliminate the pose variations.

Although rigid body parts are simple to implement, due to the inaccurate pedestrian detection and occlusion, such trivial partition is not beneficial for learning discriminative feature which should be aligned to full person. The representation based on flexible body parts improves the full person alignment to a certain extent by localizing appropriate parts. Yet, such methods usually have a much more complex pipeline or require extra prior knowledge (\eg human pose). Furthermore, the existing methods based on either rigid or flexible body parts all ignore the relationship between the body parts. Whereas, the pedestrian can always be decomposed into a sequence of body parts from head to foot. Based on this simple yet important property of pedestrian, we propose to learn all the parts together in a sequence way rather than discarding the spatial context by independent part feature extractors. For that, we naturally apply LSTM for the sequence-level modeling. The LSTM is also used in~\cite{varior2016siamese} to model the spatial contextual information between body parts. However, the authors first divide the input image into rigid parts and extract hand-craft feature for each individual part, then use LSTM to model the spatial relationship in a separate step. The proposed Deep-Person jointly integrates the deep feature extraction and sequence modeling in an end-to-end fashion, leading to more discriminative features focusing mainly on full body for person Re-ID. 

\paragraph{Approaches based on joint multi-loss learning} Recently, Zheng~\etal~\cite{zheng2016person} propose that person Re-ID lies in between instance retrieval task and image classification task. For the first point, the person Re-ID is regarded as a ranking task, where a ranking loss is adopted for feature learning. 
%For instance, Ding~\etal~\cite{ding2015deep} use a classic triplet loss for distance measure. 
In~\cite{cheng2016person}, a new term is added to the original triplet loss to further pull the instances of the same person closer. Hermans~\etal~\cite{hermans2017defense} introduce a variant of the standard triplet loss using hard mining within the batch. From the point of view of classification task, the person Re-ID problem is usually solved with a Softmax loss. There are two ways to perform person Re-ID as a classification task. The first one is known as verification network~\cite{li2014deepreid}, which takes a pair of images as input and determines whether they belong to the same identity or not by a binary classification network. The second one is called identification network~\cite{zheng2016person}, namely multi-class recognition network, where each individual is regarded as an independent category. 

The classification task and ranking task are complementary to each other. Some approaches optimize the network simultaneously with both type of loss functions, leveraging the complementary advantages of these two tasks. For example, In~\cite{chen2017multi,wang2016joint}, triplet loss and verification loss are trained together. The identification loss and verification loss are simultaneously optimized in~\cite{geng2016deep,qian2017multi}. The combination of triplet loss and identification loss is adopted in~\cite{liu2017end} to optimize the comparative attention network. The proposed Deep-Person also rely on triplet loss and identification loss. Different from~\cite{liu2017end}, the identification loss on the novel part-based representation is also adopted in addition to the identification loss on global representation, further boosting the discriminative ability of learned feature for person Re-ID. 

\paragraph{Sequence modeling using LSTM}
The LSTM has been widely used in many sequence-based problems, such as image caption~\cite{xu2015show}, machine translation~\cite{sutskever2014sequence}, speech recognition~\cite{graves2013speech}, text recognition~\cite{shi2017end} etc.
The LSTM has also shown a high potential in image classification~\cite{Wang2016unified} and object detection~\cite{bell2016inside}, where LSTM models the spatial dependencies and captures richer contextual information. Unlike these typical CNN-RNN frameworks that use an LSTM to model the context and learn a better representation directly, the LSTM in our network is only used to describe a person structure as a sequence. The backbone CNN features (implicitly integrating both global and part information) are our final representation for person Re-ID, which is influenced by the branch of LSTM through backward procedure. In this sense, the branch for the LSTM in our network can be considered as a special “loss function” mainly designed for parts, which is complementary to the two other loss functions: Softmax and Triplet loss. This usage of LSTM has never been seen in the previous CNN-RNN approaches.

\section{Architecture of Deep-Person} \label{sec:method}
\subsection{Overview of Deep-Person} \label{ssec:overall framework}

The proposed Deep-Person model focuses on both feature representation and feature learning. It is built upon two kinds of complementary designs detailed in the following: 1) Global representation and part-based local representation; 2) Softmax-based identification branch and ranking branch with the triplet loss.

Recent advances in person re-identification rely on deep learning to learn discriminative features from detected pedestrians. As pointed out in~\cite{li2017learning}, the learned representation of the full body focuses more on global information such as shape. Whereas, in some cases, only certain body parts such as the head, upper body, or lower body are discriminative for person re-identification~\cite{cheng2016person,shi2016embedding,varior2016siamese}. In this sense, the part-based local representation of detected pedestrian is complementary to the global representation. We propose to use the LSTM-based RNN to naturally model the spatial dependency between each part of the pedestrian body, which can be seen as a sequence of body parts from head to foot. Hopefully, combining the complementary global representation and the LSTM-based local representation would strengthen the discriminative ability of learned features for person re-identification.  

Recently, a common component of most deep models for person Re-ID is the Softmax-based identification branch that distinguishes different IDs based on the learned deep features. Yet, the training objective for identification branch is not totally consistent with the goal of person Re-ID, which aims to pair each probe with one of the gallery images. This is because that the identification branch does not explicitly learn a similarity measurement which is required for person Re-ID. Recently, as advocated in~\cite{liu2017end}, a distance ranking branch with the triplet loss helps to learn a similarity measure between a pair of images. In this sense, the identification branch and the distance ranking branch with the triplet loss constitute another complementarity. 

Our proposed Deep-Person model leverages the above two kinds of complementaries. The overall architecture is depicted in Fig.~\ref{fig:pipeline}. It is composed of two main components: (1) Backbone network for learning shared low-level feature $f_b$ with size $H \times W \times C$; (2) Multi-branch network to learn a highly discriminative pedestrian representation thanks to the three complementary branches: part-based identification branch $B_p^I$, global-based identification branch $B_g^I$, and distance ranking branch $B_t^R$ using the triplet loss. A joint learning strategy is adopted for simultaneously optimizing per-branch feature representation and discovering correlated complementary information.

\begin{figure}[t]
\begin{center}
   \includegraphics[width=0.7\linewidth]{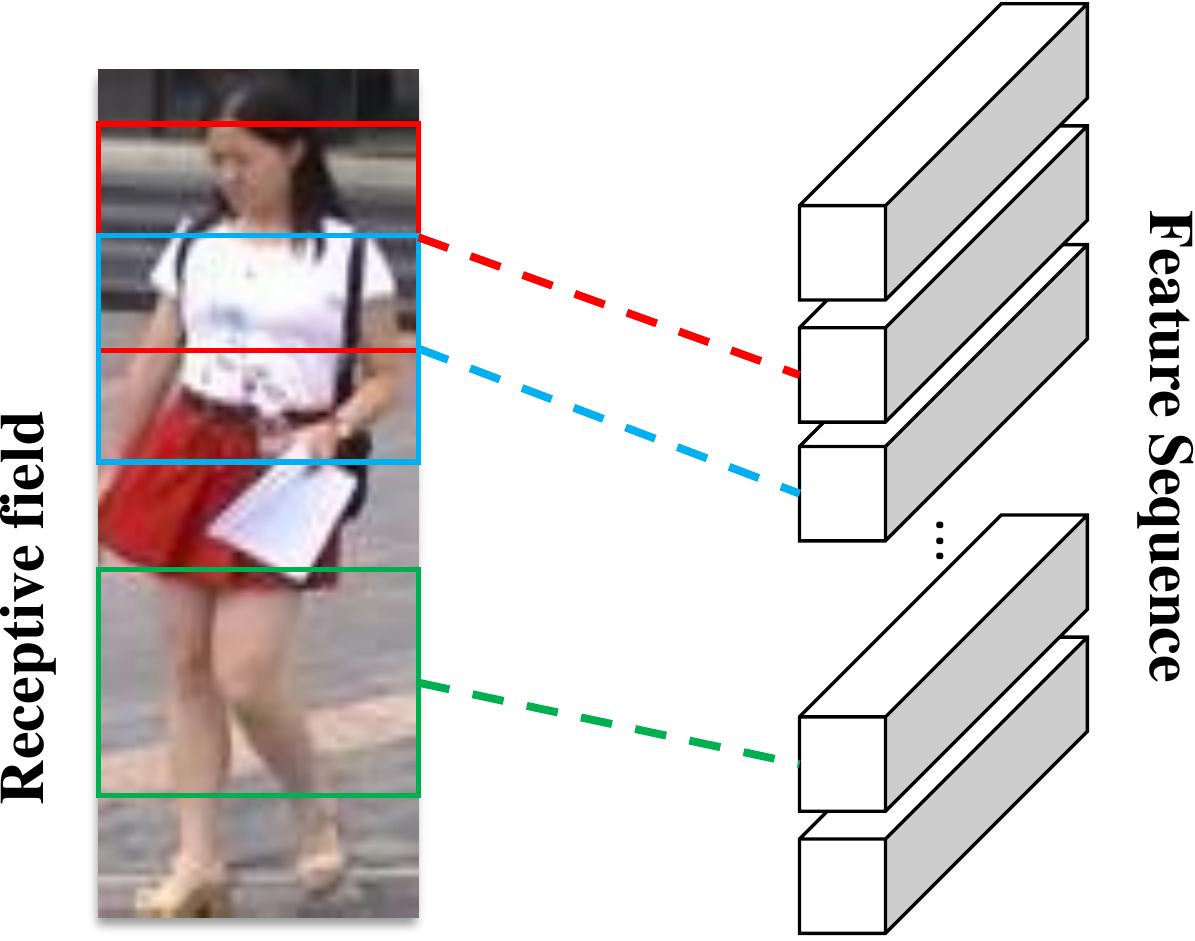}
\end{center}
   \caption{Each vector in the feature sequence describes the region of the corresponding receptive field in the raw image.}
\label{fig:receptivefield}
\end{figure}

\subsection{Part-Sequence Learning Using LSTM} \label{subsec:partlstm}

It has been shown in many methods that part-based representation learns feature focusing on some person details, which is useful for person Re-ID. Most part-based methods roughly decompose the extracted pedestrian into predefined rigid body parts which approximately correspond to head, shoulder, upper-body, upper-leg and lower-leg, respectively. Each segmented part is then fed into an individual branch to learn the corresponding local feature. This may give interesting results in some cases. Yet, the individual process of each part ignores the spatial dependencies between different parts, which is useful to learn discriminative and robust features focusing mainly on the whole body for person Re-ID. 
%Besides, such a partition exhibits a potential misalignment between different images of the same identity due to pose variations, inaccurate pedestrian detection and/or occlusion. Some examples are given in Fig.~\ref{fig:misalignment}. The upper body of the left image in Fig.~\ref{fig:misalignment}(a) is aligned with the head part of the right image.

\begin{figure}[t]
\begin{center}
   \includegraphics[width=0.7\linewidth]{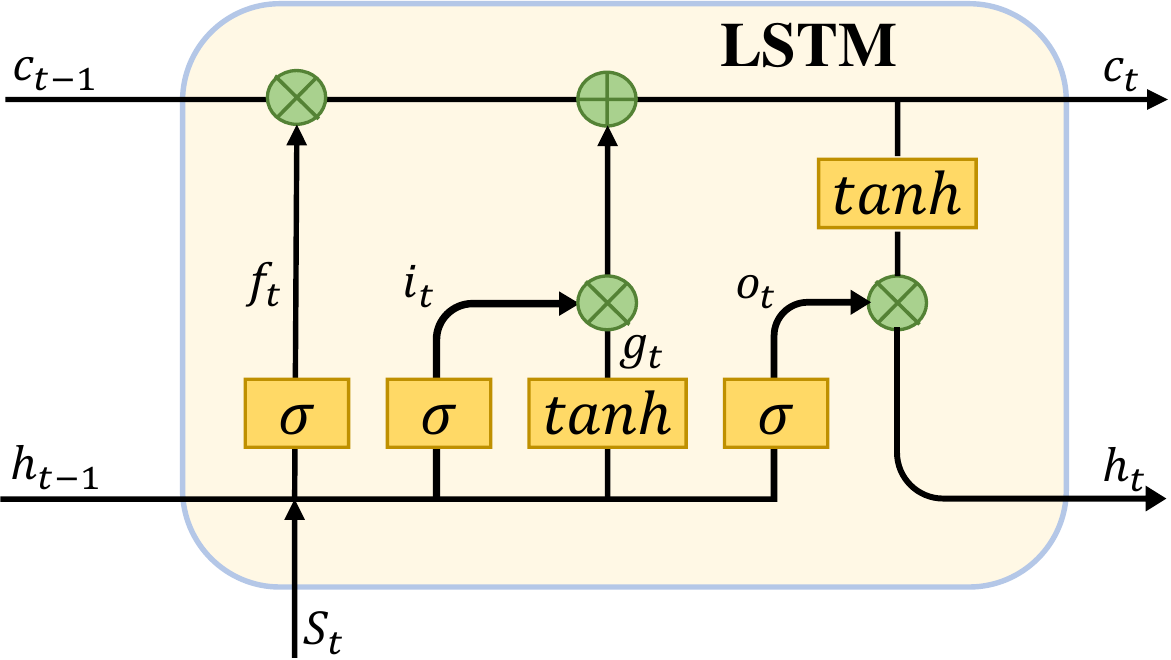}
\end{center}
   \caption{The structure of LSTM unit. An LSTM consists of a cell module $c_t$ and three gates, namely the input gate $i_t$, the output gate $o_t$ and the forget gate $f_t$. At the time step $t$, LSTM takes the $t$-th slice feature $S_t$ as well as the previous hidden state $h_{t-1}$ as inputs, and predicts a feature vector $O_t$ (generated by attaching a fully-connected layer after $h_t$). }
\label{fig:lstm_unit}
\end{figure}

We notice that pedestrian in images can be decomposed into a sequence of body parts from head to foot. Even though each part does not always lie in the same position in different images, all the pedestrian parts can be modeled in a sequence way thanks to the priori knowledge of body structure. The sequential representation of pedestrian naturally motivates us to resort to RNN with LSTM cells, which is detailed in Fig.~\ref{fig:lstm_unit}. The recurrent connections between the LSTM units enable the RNN to yield features based on the historical inputs. Therefore the learned features at any point are refined by the spatial contexts. What's more, benefiting from the internal gating mechanisms, LSTM can control the information flow from the current state to the next state. Consequently, LSTM cell is competent to propagate certain relevant contexts and filter out some irrelevant parts. Based on the above insights, we propose to adopt LSTM to model the sequence of body parts for person Re-ID. 

More specifically, to get the spatial contexts, we extract directly a sequence of feature vectors from the shared low-level features $f_{b}$ without any explicit segmentation. As depicted in Fig.~\ref{fig:pipeline}, each row of $f_b$ undergoes an average pooling ({\em i.e.}, average pooling with $1 \times W$ kernel), which results in a corresponding feature vector of length $C$. As illustrated in Fig.~\ref{fig:receptivefield}, each feature vector describes a rectangle region in raw image given by the corresponding receptive field. A two-layer Bidirectional LSTM is then built upon the feature sequence. Thanks to the context modeling of LSTM, each resulting feature vector of length $U$ may describe better its associated part. Finally, all the resulting feature vectors characterizing the underlying local parts are concatenated together as the final part-based person representation. This part-based feature is learned via a Softmax layer with $N_c$ output neurons, where $N_c$ denotes the number of pedestrian identities.

\subsection{Global Representation Learning} \label{subsec:global}
Part-based features focus more on discriminative pedestrian details. It is difficult to distinguish two different identities with very similar visual details ({\em e.g.}, wearing the same clothes) using only part-based representation. 
In this case, the shape information is required to distinguish them. Indeed, the global feature is complementary to the part-based local feature, and contains more high-level semantics, like shape. Similar to many deep neural networks for person Re-ID, we straightforward extract a global representation by inserting a global average pooling and a fully-connected layer after the shared low-level feature $f_{b}$. A Softmax layer with $N_c$ output neurons is then appended for global feature learning.

\subsection{Deep Metric Learning with Triplet Loss} \label{sssec:method_tripletloss}

The part-based and global representation learning described in Sec.~\ref{subsec:partlstm} and~\ref{subsec:global} do not explicitly learn a similarity measure which is required for person Re-ID during test. We propose a third branch in Deep-Person model that is responsible for distance ranking. For that, we apply another independent global average pooling to the shared low-level feature $f_b$, which results in a feature $f_m$ in the metric space for similarity estimation. This feature $f_m$ is also adopted as the final pedestrian descriptor for person Re-ID. 

More specifically, we adopt the improved triplet loss for deep metric learning proposed in~\cite{hermans2017defense}. The main idea lies in forming batches named $PK sampling$ by randomly sampling $P$ classes (person identities), and then randomly sampling $K$ images of each class (person), thus resulting in a batch of $PK$ images. Given such a mini-batch $X=\{x_i\}_{i=1}^{PK}$, we can constitute a triplet $T_i=\{x_i^o, x_i^+, x_i^-\}$ for each selected anchor image $x_i$, where $x_i^+$ ({\em resp.} $x_i^-$) denotes a positive ({\em resp.} negative) sample in the mini-batch having the same ({\em resp.} different) person ID as $x_i^o$. Since hard triplet mining strategy is crucial for learning based on triplet loss, only the hardest positive and negative sample in the mini-batch are selected for each anchor sample to form the triplets for loss computation: 
\begin{equation} \label{eq:triplet_loss}
\begin{split}
{\cal L}_{trp} = \frac{1}{PK}\sum_{i=1}^{PK}[m + \max \limits_+ D(f_m(x_i^o), f_m(x_i^+)) \\ 
- \min \limits_- D(f_m(x_i^o), f_m(x_i^-))]_+,
\end{split}
\end{equation}
where $[\cdot]_+ = \max (\cdot, 0)$, $m$ is a margin that is enforced between positive and negative pairs, and $D(a, b)$ is referred as the distance function between feature vector $a$ and $b$. In this paper, we use the Euclidean distance $D(f_m(x_i), f_m(x_j)) = \|f_m(x_i) - f_m(x_j)\|_2$ as the distance metric. This objective function encourages the features of positive pairs to be closer in the learned feature space than the negative pairs by a predefined margin $m$.

\subsection{Model Training}

As discussed in Sec.~\ref{ssec:overall framework}, different branches have complementary strengths for learning discriminative pedestrian descriptors. To leverage these complementaries, we jointly train the whole network to predict person identity for both part-based and global feature learning while also satisfying the triplet objective. There are two identification subnets built with Softmax loss for multi-class person identification task, defined as:
\begin{equation} \label{eq:cls_loss}
{\cal L}_{cls} = -\frac{1}{N}\sum_{i=1}^{N} \log \frac{\exp(W_{y_i}^T f_i+b_{y_i})}{\sum_{j=1}^{N_c} \exp(W_j^T f_i+b_j)},
\end{equation}
where $f_i$ is the classification feature of $i$-th sample, $y_i$ is the identity of $i$-th sample, $N$ is the number of samples, $W_j$ and $b_j$ is respectively the weight and bias of the classifier for $j$-th identity. To simplify the notation and without any ambiguity, we replace $f_i$ with $f_p$ ({\em resp.} $f_g$) for the corresponding part-based ({\em resp.} global-based) classification loss function ${\cal L}_{cls\_p}$ ({\em resp.} ${\cal L}_{cls\_g}$). Then the final loss is given by:
\begin{equation} \label{eq:all_loss}
{\cal L} = \lambda_1 {\cal L}_{trp} + \lambda_2 {\cal L}_{cls\_p} + \lambda_3 {\cal L}_{cls\_g},
\end{equation}
where $\lambda_i$ ($i=1,2,3$) denotes the loss weight for different branch.

\section{Experiments} \label{sec:experiments}

\subsection{Implementation Details} \label{subsec:implementation}

The proposed Deep-Person model is built on the PyTorch framework. The backbone network is the ResNet-50~\cite{he2016deep} model pre-trained on ImageNet, where the global average pooling and fully connected layers are discarded. The parameter $C$, number of channels of $f_b$ (see Fig.~\ref{fig:pipeline}) is set to 2048, and $U$, the number of hidden units in BLSTM is set to 256. Depending on the height of input images and the backbone network, $H$ is set to 8 in this paper.

We follow common dataset augmentation strategies with different scales and aspect ratios to train the Deep-Person model. Concretely, all training images are first resized to 256$\times$128. Then, we randomly crop each resized image with scale in the interval [0.64, 1.0] and aspect ratio in [2, 3]. Finally, the cropped patch is resized again to 256$\times$128. Randomly horizontal flip with a probability of 0.5 is also applied. During testing phase, images are simply resized to 256$\times$128. 
The globally pooled feature $f_m$, input of the distance ranking branch is used as the pedestrian descriptor for retrieval. Before feeding the input image to the network, we follow the image normalization procedure in original ResNet-50 model trained on ImageNet by subtracting the mean value and then dividing by the standard deviation.

As described in~\ref{sssec:method_tripletloss}, $PK sampling$ is used to form a batch, where $P$ and $K$ denote respectively the number of sampled classes and instances of each class. To train our Deep-Person model, we redefine the notion of ``epoch'' such that every $[N_c/P]$ trained batches form an ``epoch''. This redefinition implies that an ``epoch'' covers approximately all the identities.  Jointly training such a network is not trivial in the beginning of training stages. We apply gradient clipping to avoid gradient explosion. The Adam with the default hyper-parameters ($\epsilon=10^{-8}$, $\beta_1=0.9$, $\beta_2=0.999$) is used to minimize the loss function ${\cal L}$ given in Eq.~\eqref{eq:all_loss}, where $\lambda_1$, $\lambda_2$, and $\lambda_3$ are all set to 1. The margin $m$ in Eq.~\eqref{eq:triplet_loss} is set to 0.5. Following the common practice of learning rate decaying schedule in~\cite{hermans2017defense}, we set the initial learning rate to $3\times10^{-4}$ and apply the decay schedule at epoch 100. The total number of training epochs for all conducted experiments are set to 150. The mini-batch size $PK$ is set to 128 for all experiments. Following different average number of samples for each identity in different datasets detailed in Sec.~\ref{subsec:datasets}, $K$ is set respectively to 8, 4, and 8 for Market-1501, CHUK03, and DukeMTMC-reID.

\subsection{Datasets and Evaluation Protocol} \label{subsec:datasets}
\paragraph{Datasets} We evaluate our proposed method, Deep-Person, on three widely used large-scale datasets:  Market-1501~\cite{zheng2015scalable}, CUHK03~\cite{li2014deepreid}, and DukeMTMC-reID~\cite{zheng2017unlabeled,ristani2016performance} dataset. A brief description of them is given as follows:

\textit{Market-1501}: This dataset consists of 32,668 images of 1,501 identities captured by six cameras in front of a supermarket. The provided pedestrian bounding boxes are given by Deformable Part Model (DPM)~\cite{felzenszwalb2008discriminatively} and further validated by manual annotations. The dataset is divided into training set consisting of 12,936 images of 751 identities and testing dataset containing the other 19,732 images of 750 identities. For each identity in the testing set, only one image from each camera is selected as a query, resulting in 3,368 query images in total.   

\textit{CUHK03}: This dataset contains 14,097 images of 1,467 identities shot by six cameras in the CUHK campus. It provides two types of annotations: manually labeled pedestrian bounding boxes and automatic detections given by DPM detector. We conduct experiments on both types of annotated datasets named as {\em labeled} dataset and {\em detected} dataset, respectively. This dataset also offers 20 random splits, each of which selects 100 test identities and use the other 1,367 persons for training. The average performance on all splits is reported for evaluation on this dataset.

\textit{DukeMTMC-reID}: This dataset is a subset of Duke-MTMC~\cite{ristani2016performance} for image-based re-identification, in the format of the Market-1501 dataset. It is composed of 36,411 images of 1,812 different identities taken by 8 high-resolution cameras, where 1,404 identities appear in more than two cameras and the other 408 identities are regarded as distractor IDs. Among the 1404 identities,  16,522 images of 702 identities are used for training, the other 702 identities are divided into 2,228 query images and 17,661 gallery images.

\paragraph{Evaluation Protocol} We follow the standard evaluation protocol. Concretely, the precisions of rank-1, rank-5, and rank-10 are reported for CUHK03. The cumulative matching characteristics (CMC) at rank-1 and mean average precision (mAP) for performance evaluation on Market-1501 and DukeMTMC-reID. Following most related works, the evaluation on CHUK03 and DukeMTMC-reID is performed under single query setting. Both single and multiple query settings are used for Market-1501 dataset.

\subsection{Comparison with Related Methods}
We compare the proposed method, Deep-Person, with the state-of-the-art approaches on Market-1501, CUHK03, and DukeMTMC-reID datasets. The proposed Deep-Person consistently outperforms the state-of-the-art methods on all three datasets. The details are given as follows:

%----------------------------------
\setlength{\tabcolsep}{16pt}
\begin{table}[tb]
\centering
\caption{Quantitative comparison with state-of-the-art methods on DukeMTMC-reID dataset.}
\vspace{+1ex}
\begin{tabular}{l|cc}
\hline
Methods & rank1 & mAP \\
\hline
BoW+KISSME~\cite{zheng2015scalable} & 25.13 & 12.17 \\
LOMO+XQDA~\cite{liao2015person} & 30.75 & 17.04\\
LSRO~\cite{zheng2017unlabeled} & 67.68  & 47.13\\
AttIDNet~\cite{lin2017improving} & 70.69  & 51.88\\
PAN~\cite{zheng2017pedestrian}& 71.59 & 51.51 \\
ACRN~\cite{schumann2017person} & 72.58 & 51.96\\
SVDNet~\cite{sun2017svdnet} & 76.70  & 56.80\\
DPFL~\cite{chen2017person} & \textbf{\color{blue}{79.20}} & \textbf{\color{blue}{60.60}} \\
\hline
\textbf{Deep-Person (Ours)} &\textbf{\color{red}{80.90}}& \textbf{\color{red}{64.80}} \\
\hline
\end{tabular}
\label{table:DukeMTMC_art}
\end{table}
%-----------------------------------

%-------------------------------
\setlength{\tabcolsep}{10pt}
\begin{table}[tb]
\centering
\caption{Comparison with state-of-the-art results on Market-1501. The $1^{st}$/$2^{nd}$ best result is highlighted in red/blue.}
\vspace{+1ex}
\begin{tabular}{l|*{2}{p{0.9cm}<{\centering}}|*{2}{p{0.9cm}<{\centering}}}
\hline
\multirow{2}{*}{Methods}   & \multicolumn{2}{c|}{Single Query} & \multicolumn{2}{c}{Multiple Query}\\
\cline{2-5}
 & rank1 & mAP & rank1 & mAP   \\
\hline
OL-MANS~\cite{zhou2017efficient} & 60.67 & - & 66.80 & - \\
DNS~\cite{zhang2016learning} & 61.02 & 35.68 & 71.56 & 46.03 \\
Gated S-CNN~\cite{varior2016gated} & 65.88 & 39.55 & 76.04 & 48.45 \\
CRAFT-MFA~\cite{chen2017personPAMI} & 68.70 & 42.30 & 77.70 & 50.30 \\
P2S~\cite{zhou2017point} & 70.72 & 44.27 & 85.78 & 55.73 \\
CADL~\cite{lin2017consistent} & 73.84 & 47.11 & 80.85 & 55.58 \\
Spindle~\cite{zhao2017spindle} & 76.90 & - & - & - \\
MSCAN~\cite{li2017learning} & 80.31 & 57.53 & 86.79 & 66.70 \\
SVDNet~\cite{sun2017svdnet} & 82.30 & 62.10 & - & - \\
Part-Aligned~\cite{zhao2017deeply} & 81.00 & 63.40 & - & - \\
PDC~\cite{su2017pose} & 84.14 & 63.41 & - & - \\
JLML~\cite{li2017JLML} & 85.10 & 65.50 & 89.70 & 74.50 \\
LSRO~\cite{zheng2017unlabeled} & 83.97 & 66.07 & 88.42 & 76.10 \\
SSM~\cite{bai2017scalable} & 82.21 & 68.80 & 88.18 & 76.18 \\
TriNet~\cite{hermans2017defense} & 84.92 & 69.14 & 90.53 & 76.42 \\
DPFL~\cite{chen2017person} &\textbf{\color{blue}{88.60}} &\textbf{\color{blue}{72.60}} &\textbf{\color{blue}{92.20}} &\textbf{\color{blue}{80.40}} \\
\hline
\textbf{Deep-Person (Ours)} & \textbf{\color{red}{92.31}} & \textbf{\color{red}{79.58}} & \textbf{\color{red}{94.48}} & \textbf{\color{red}{85.09}} \\
%\textbf{Deep-Person (Re-ra.)} & \textbf{93.71} & \textbf{90.84} & \textbf{95.16} & \textbf{93.67} \\
\hline
\end{tabular}
\label{table:market1501_art}
\end{table}
%-------------------------------

% ---------------------
\begin{table}[!t]
\centering
\caption{Evaluation on CUHK03 in terms of rank-1 (r1), rank-5 (r5), and rank-10 (r10) matching rate, using manually labeled pedestrian bounding boxes and automatic detections by DPM.}
\vspace{+1ex}
\begin{tabular}{l|*{3}{p{0.5cm}<{\centering}}|*{3}{p{0.5cm}<{\centering}}}
%\hline
\hline
\multirow{2}{*}{Methods}   & \multicolumn{3}{c|}{Labeled} & \multicolumn{3}{c}{Detected}\\ \cline{2-7}
& r1 & r5 & r10  & r1 & r5 & r10 \\ \hline
EDM\cite{shi2016embedding} & 61.3	& 88.9 & 96.4	& 52.1 & 82.9 & 91.8 \\
OL-MANS~\cite{zhou2017efficient} & 61.7 & 88.4 & 95.2 & 62.7 & 87.6 & 93.8 \\
DNS~\cite{zhang2016learning} & 62.5 & 90.0 & 94.8 &	54.7 &	84.7 &	94.8 \\
Gated S-CNN~\cite{varior2016gated} & - & - & - & 68.1 & 88.1 & 94.6 \\
GOG~\cite{matsukawa2016hierarchical} & 67.3 & 91.0 & 96.0 & 65.5 & 88.4 & 93.7 \\
DictRW~\cite{cheng2017DictRW} & 71.1 & 91.7 & 94.7 & - & - & - \\
MSCAN~\cite{li2017learning} & 74.2 & 94.3 & 97.5 & 68.0 & 91.0 & 95.4 \\
MTDnet~\cite{chen2017multi} & 74.7 & 96.0 & 97.5 & - & - & - \\
Quadruplet~\cite{chen2017beyond} & 75.5 & 95.2 & 99.2 & - & - & - \\
SSM~\cite{bai2017scalable} & 76.6 & 94.6 & 98.0 & 72.7 & 92.4 & 96.1 \\
MuDeep~\cite{qian2017multi} & 76.9 & 96.1 & 98.4 & 75.6 & 94.4 & 97.5 \\
SVDNet~\cite{sun2017svdnet} & - & - & - & 81.8 & 95.2 & 97.2 \\
CRAFT-MFA~\cite{chen2017personPAMI} & - & - & - & 84.3 & 97.1 & 98.3 \\
LSRO~\cite{zheng2017unlabeled} & - & - & - & \textbf{\color{blue}{84.6}} & \textbf{\color{blue}{97.6}} & \textbf{\color{blue}{98.9}} \\
JLML~\cite{li2017JLML} & 83.2 & 98.0 & \textbf{\color{blue}{99.4}} & 80.6 & 96.9 & 98.7 \\
Part-Aligned~\cite{zhao2017deeply} & 85.4 & 97.6 & \textbf{\color{blue}{99.4}} & 81.6 & 97.3 & 98.4 \\
Spindle~\cite{zhao2017spindle} & 88.5 & 97.8 & 98.6 & - & - & - \\
PDC~\cite{su2017pose} & \textbf{\color{blue}{88.7}} & \textbf{\color{blue}{98.6}} & 99.2 & 78.3 & 94.8 & 97.2 \\
\hline
\textbf{Deep-Person (Ours)} & \textbf{\color{red}{91.5}} & \textbf{\color{red}{99.0}} & \textbf{\color{red}{99.5}} & \textbf{\color{red}{89.4}} & \textbf{\color{red}{98.2}} & \textbf{\color{red}{99.1}}\\
%\hline
\hline
\end{tabular}
\label{table:CUHK03_art}
\end{table}
%----------------------------------

\paragraph{Evaluation on DukeMTMC-reID} The comparison with state-of-the-art methods on DukeMTMC-reID dataset is depicted in Table~\ref{table:DukeMTMC_art}. 
Our Deep-Person also outperforms all the state-of-the-art approaches, achieving an improvement of 1.7\% rank-1 accuracy and 4.2\% mAP. It is worth noting that the previous state-of-the-art DPFL~\cite{chen2017person} take multi-scale person images as input. With this design, the performance is expected to be further improved.

\paragraph{Evaluation on Market-1501} 
As depicted in Table~\ref{table:market1501_art}, the proposed Deep-Person achieves compelling results on Market-1501 dataset. Concretely, our Deep-Person significantly improves the state-of-the-art results by about \textbf{7\%} on mAP and 3.7\% on rank-1 matching rate under single query mode. When combining with an effective re-ranking approach~\cite{zhong2017re}, the performance is further boosted, reaching \textbf{90.84\%} mAP. To the best of our knowledge, this is the first time that an mAP higher than 90\% is achieved on the Market-1501 dataset with single query setting. We also observe similar performance improvements using multiple query setting on this dataset, getting 4.7\% and 2.3\% improvement on mAP and rank-1, respectively.  

\paragraph{Evaluation on CUHK03}
The evaluation of Deep-Person on CUHK03 dataset in terms of rank-1, rank-5, and rank-10 matching rate is given in Table~\ref{table:CUHK03_art}. Using the manually annotated pedestrian bounding boxes, our Deep-Person yields 2.8\% rank-1 accuracy improvement. An improvement of 4.8\% rank-1 accuracy is achieved when using an automatic method DPM to extract pedestrians. The later setting is coherent with practical application, which demonstrates the potential and robustness of Deep-Person in practice. Deep-Person also outperforms the state-of-the-art methods under both rank-5 and rank-10 matching rate.

%-----------------------------------
\renewcommand\arraystretch{1.2}
\setlength{\tabcolsep}{16pt}
\begin{table}[tb]
% \small
\centering
\caption{Effectiveness of the complementary advantage between global features and the novel LSTM-based part features.}
\vspace{+1ex}
\begin{tabular}{l|cc}
\hline
\multirow{2}{*}{Feature type} & \multicolumn{2}{c}{Single Query} \\ \cline{2-3} 
& rank1 & mAP \\ \hline
$B_g^I$ &85.33 &64.71 \\ 
$B_p^I$ w/o LSTM &82.39 &60.75 \\ 
$B_g^I$ + $B_p^I$ w/o LSTM &86.49 &66.74 \\ 
$B_g^I$ + $B_p^I$ &\textbf{\color{red}{87.74}} &\textbf{\color{red}{69.82}} \\ \hline
\end{tabular}
\label{table:ablation_study_lstm}
\end{table}
%-----------------------------

\subsection{Ablation Study}
We further evaluate several variants of Deep-Person to verify the effectiveness of each individual component. Without loss of generality, the ablation study is performed on Market-1501 dataset under single query setting, using the same settings as described in Sec.~\ref{subsec:implementation}.

\paragraph{Effectiveness of LSTM-based Parts}
The proposed Deep-Person leverages the complementary information between global representation and a novel LSTM-based part representation. We evaluate the contribution of this complementary advantage and the effect of the novel LSTM-based part representation. For that, we discard the ranking branch $B_t^R$ in Deep-Person. As depicted in Table~\ref{table:ablation_study_lstm}, the combination of global and part-based branches outperforms each individual branch alone. Furthermore, the adopted LSTM achieves 3.1\% performance gain in mAP. Note that for a fair comparison between the variant $B_g^I$ + $B_p^I$ without LSTM and $B_g^I$ + $B_p^I$, a fully-connected layer with 256 neurons is attached after each part slice in the former so that the two models have approximately the same number of model parameters.

% ---------------------
\begin{figure}[t]
\begin{center}
   \includegraphics[width=\linewidth,height=0.4\textheight]{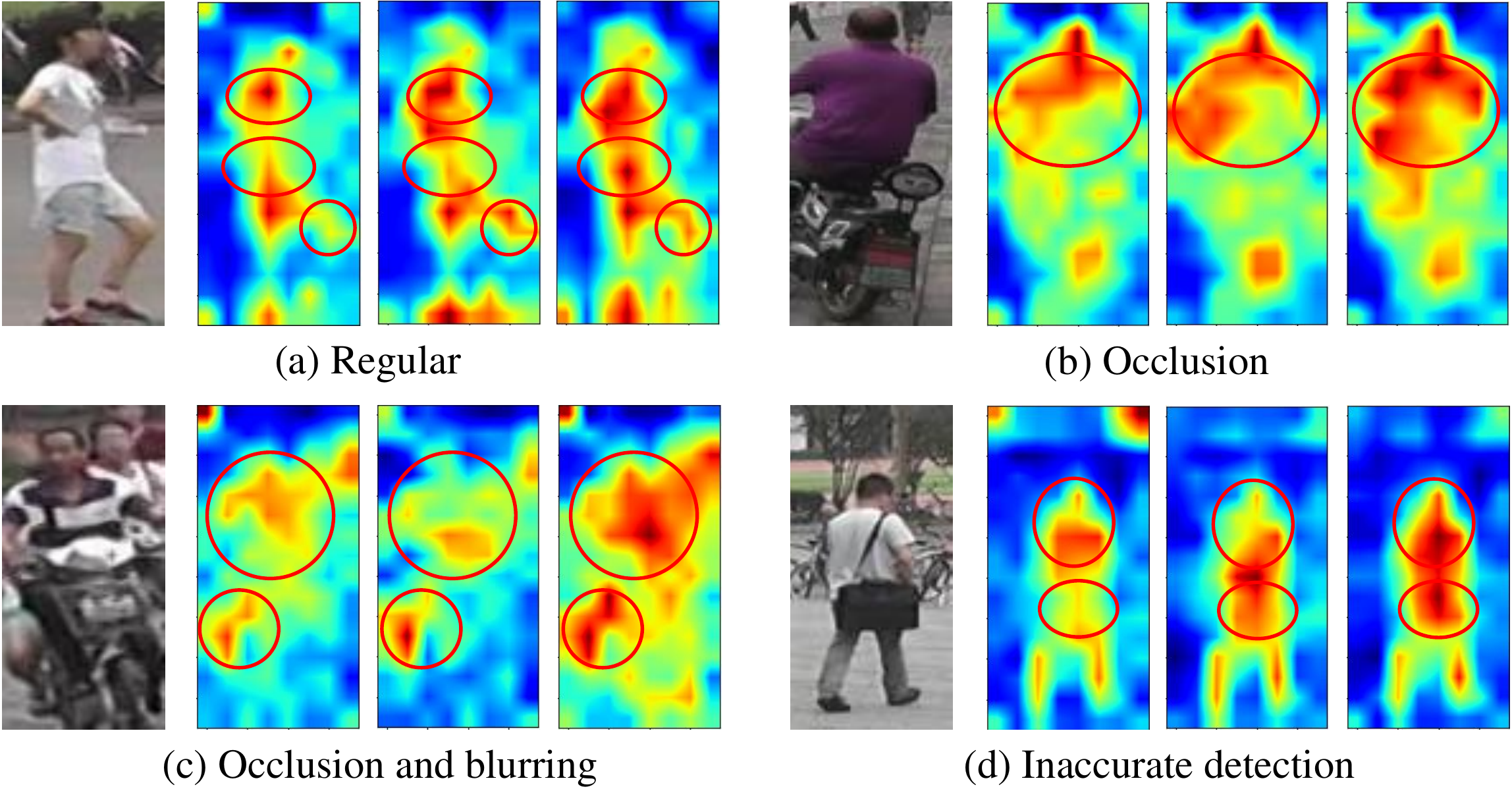}
\end{center}
   \caption{Visualization of feature maps extracted from three variants of Deep-Person. From left to right in (a-d): raw image, the feature map of using $B_g^I$ branch alone, $B_g^I$ + $B_p^I$ without LSTM, and $B_g^I$ + $B_p^I$ with LSTM, respectively. {Major differences are enclosed by red circles.}}
\label{fig:feaVis}
\end{figure}

\begin{figure}[!h]
\begin{center}
   \includegraphics[width=\linewidth,height=0.35\textheight]{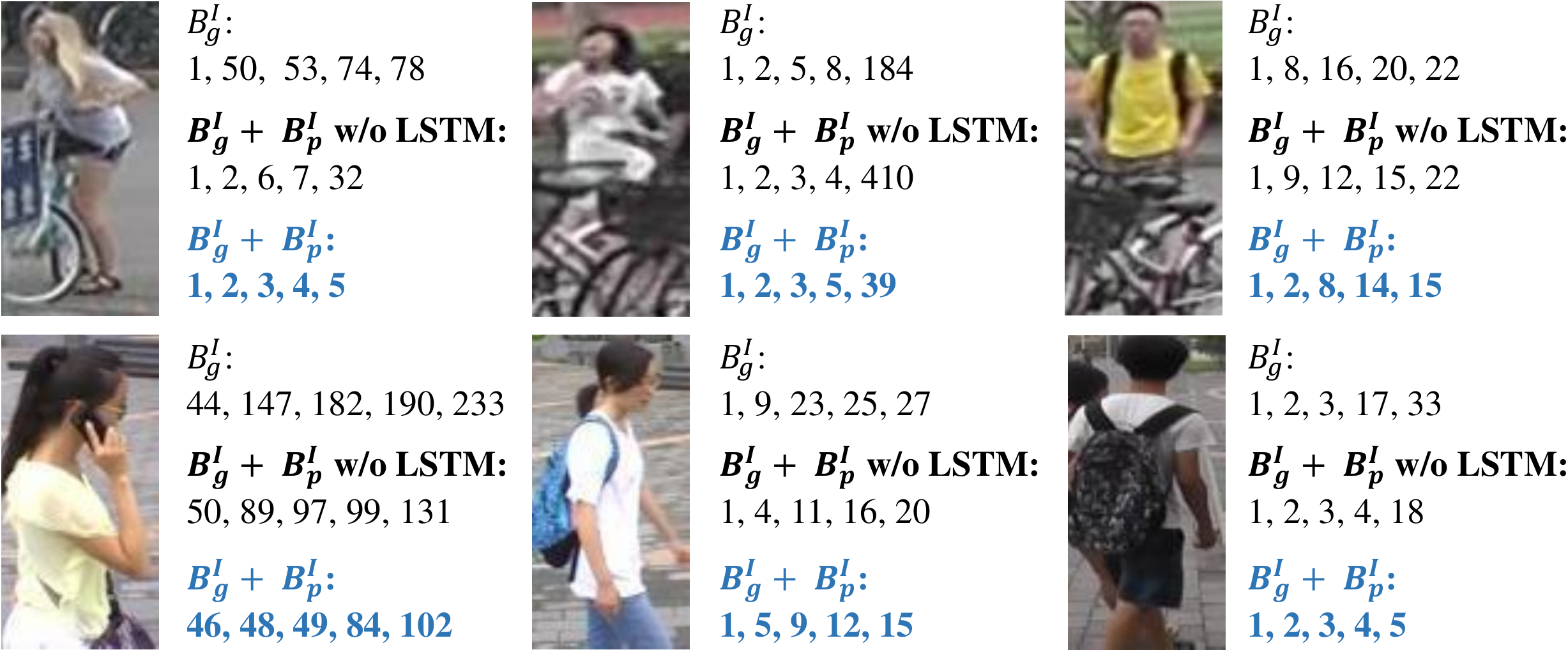}
\end{center}
   \caption{Comparison of the gallery match ranks of each probe image under occlusion, blurring, background cluster, and imperfect detection. Each probe image has multiple matches in the gallery. Smaller numbers mean better ranking performances.}
\label{fig:rankVis}
\end{figure}
%-----------------------------

To get some insights about the improvements, we analyze the learned feature maps $f_b$ of three variants of Deep-Person: $B_g^I$ branch alone, $B_g^I$ + $B_p^I$ without LSTM, and  $B_g^I$ + $B_p^I$ with LSTM. Some illustrations are given in Fig.~\ref{fig:feaVis}. Using global branch $B_g^I$ alone only captures coarse areas with few part details. However, the missed part details such as the upper-body or backpack are potentially meaningful to identify the person. Combining with part-based representation without LSTM enriches some fine details. Adopting LSTM in the combination of global and part-based branches focuses more on the pedestrian with fine details, ignoring irrelevant regions (\eg the top corners in Fig.~\ref{fig:feaVis} (d) and the motorbike in Fig.~\ref{fig:feaVis} (b)). As a result, the learned features of $B_g^I$ + $B_p^I$ using LSTM align better to the full person and thus are more complete and robust for person Re-ID. The above analyses are consistent with the gallery match ranks test illustrated in Fig.~\ref{fig:rankVis}. From this figure, one can see that $B_g^I$ + $B_p^I$ with LSTM achieves better matching ranks than the other two variants in most cases. This implies that LSTM-based part representation with global descriptor is capable of mitigating occlusion, blurring, background cluster and imperfect detection. 
% making the learned features better align to the whole person.

%-----------------------------
\newcommand{\tabincell}[2]{\begin{tabular}{@{}#1@{}}#2\end{tabular}}
\renewcommand\arraystretch{1}
\setlength{\tabcolsep}{11pt}
\begin{table}[tb]
\centering
\caption{Comparison with state-of-the-art part-based models on Market-1501 dataset.}
\vspace{+1ex}
\begin{tabular}{l|l|cc}
\hline
\multirow{2}{*}{Category} & \multirow{2}{*}{Methods} & \multicolumn{2}{c}{Single Query} \\ \cline{3-4} 
& & rank1 & mAP \\ \hline
\multirow{4}{*}{\tabincell{c}{Rigid\\Partition}} & SCSP~\cite{chen2016similarity} & 51.90 & 26.35 \\
& LSTM-Siamese~\cite{varior2016siamese} & 61.60 & 35.31 \\
&MR B-CNN~\cite{ustinova2017multi} & 66.36 & 41.17 \\ 
&JLML~\cite{li2017JLML} & 85.10 & 65.50 \\ \hline
\multirow{4}{*}{\tabincell{c}{Flexible\\Partition}} &Spindle~\cite{zhao2017spindle} & 76.90 &- \\ 
&MSCAN~\cite{li2017learning} & 80.31 & 57.53 \\ 
&Part-Aligned~\cite{zhao2017deeply} & 81.00 & 63.40 \\ 
&PDC~\cite{su2017pose} & \textbf{\color{blue}{84.14}}  & \textbf{\color{blue}{63.41}} \\  \hline
% &Part-Loss~\cite{yao2017deep} & \textbf{\color{blue}{85.60}} & \textbf{\color{blue}{65.91}} \\ \hline
\rule{0pt}{10pt}& $B_g^I$ + $B_p^I$ & \textbf{\color{red}{87.74}} & \textbf{\color{red}{69.82}} \\ 
\hline
\end{tabular}
\label{table:market1501_part_art}
\end{table}
%-------------------------------

We also compare a variant of Deep-Person using global and LSTM-based part branches with some state-of-the-art part-based models. The comparison is depicted in Table~\ref{table:market1501_part_art}. Thanks to the use of LSTM modeling spatial context information, an improvement of 6.41\% mAP and 3.6\% rank1 are achieved compared to the best part-based method. It is worth noting that Spindle~\cite{zhao2017spindle} and PDC~\cite{su2017pose} use extra pose annotations for body-part detection. As compared to the method~\cite{varior2016siamese} using also LSTM to model spatial relationship between body parts, using LSTM in an end-to-end way in Deep-Person achieves a significant improvement.

%-------------------------------
\setlength{\tabcolsep}{16pt}
\begin{table}[tb]
% \small
\centering
\caption{Effectiveness of multi-loss on the variant of Deep-Person discarding part-based branch $B_p^I$.}
\vspace{+1ex}
\begin{tabular}{l|cc}
\hline
\multirow{2}{*}{Loss type} & \multicolumn{2}{c}{Single Query} \\ \cline{2-3} 
& rank1 & mAP \\ \hline
Identification loss &85.33 &64.71 \\ 
Triplet loss &81.29 &63.51 \\ 
Identification + Triplet loss & \textbf{\color{red}{88.45}} & \textbf{\color{red}{72.96}} \\ \hline
\end{tabular}
\vspace{+1ex}
\label{table:ablation_study_multiloss}
\end{table}
%-------------------------------

%-----------------------------------------
\setlength{\tabcolsep}{16pt}
\begin{table}[tb]
% \small
\centering
\caption{Experimental results using different fused features as the final pedestrian descriptor on the $B_g^I$+$B_p^I$ two-branch model.}
\vspace{+1ex}
\begin{tabular}{l|cc}
\hline
\multirow{2}{*}{Feature type} & \multicolumn{2}{c}{Single Query} \\ \cline{2-3} 
& rank1 & mAP \\ \hline
$f_c$ & 84.65 & 67.00 \\ 
\textbf{$f_m$} & \textbf{\color{red}{87.74}} & \textbf{\color{red}{69.82}} \\ \hline
\end{tabular}
\label{table:feature_type}
\end{table}
%----------------------------------------

\paragraph{Effectiveness of Multi-loss}
Using the complementary advantage of identification and ranking loss is another important aspect of our Deep-Person. To simplify the evaluation of this complementary advantage, we discard the part-based branch $B_p^I$. As shown in Table~\ref{table:ablation_study_multiloss}, combining identification loss and ranking loss (\eg triplet loss) respectively achieves a rank1 accuracy of 88.45\% and 72.96\% mAP, which performs much better than using each of them alone. 
This reveals the complementarity of identification and ranking information to learn discriminative features for person Re-ID, as well as the effectiveness of jointly optimizing.

\paragraph{Choice of final pedestrian descriptor}

As described in Sec.~\ref{sec:method}, Deep-Person uses the global average pooled feature $f_m$ of the backbone feature $f_b$ as the final pedestrian descriptor. Complementary information between global and part-based representation is implicitly integrated into $f_m$ thanks to the combination of back-propagation error differentials of the global and part-based branch during training. Following recent works~\cite{cheng2016person,li2017learning,li2017JLML} using global and part-based branches, the concatenation of the penultimate fully connected layers of these two branches denoted as $f_c$ seems to be a reasonable alternative. We use a two-branch $B_p^I$ + $B_g^I$ variant of Deep-Person to evaluate these two choices. As shown in Table~\ref{table:feature_type}, the implicit fused feature $f_m$ significantly outperforms the directly concatenated feature $f_c$. One possible reason for this is that the identification loss makes the features near the classification layer focus more on the difference of training identities. Such feature might be discriminative for identities in training images, but is not discriminative for the unseen identities during test. Whereas, the backbone feature may be more robust and generalize better to the unseen test categories. This comparison motivates us to append the ranking branch with triplet loss after $f_m$, which is considered as the final pedestrian descriptor in Deep-Person.

\section{Conclusion}

% In this paper, we introduce a novel three-branch framework named Deep-Person to learn highly discriminative deep features for person Re-ID. Different from most existing methods which either focus on feature representation or feature learning alone, complementary advantages on both aspects are considered in Deep-Person. Concretely, local body-part and global full-body features are jointly employed. The identification loss and ranking loss are applied to simultaneously learn an ID-discriminative embedding and a similarity measurement. Furthermore, in contrast to the existing part-based methods which usually discard the spatial context of body structure, we use LSTM to enhance the discriminative ability of part representation with contextual information, leading to learned feature better aligned to full person. Extensive evaluations on three popular and challenging datasets demonstrate the superiority of the proposed Deep-Person over the state-of-the-art methods. In the future, we would like to combine the spatial context modeling with attention mechanism to automatically select more discriminative parts for person Re-ID.

In this paper, we introduce a novel three-branch framework named Deep-Person to learn highly discriminative deep features for person Re-ID. Different from most existing methods which either focus on feature representation or feature learning alone, complementary advantages on both aspects are considered in Deep-Person. Concretely, local body-part and global full-body features are jointly employed. The identification loss and ranking loss are applied to simultaneously learn an ID-discriminative embedding and a similarity measurement. In addition, in contrast to the existing part-based methods which usually discard the spatial context of body structure, we innovatively propose to regard the pedestrian as a sequence of body parts from head to foot, and apply LSTM in an end-to-end fashion to take into account the contextual information between body parts, enhancing the discriminative capacity of local feature which aligns better to full person. Such a descriptor is heuristic for person representation. Furthermore, to our best knowledge, this work is the first method to use LSTM for person Re-ID in an end-to-end way. Extensive evaluations on three popular and challenging datasets demonstrate the superiority of the proposed Deep-Person over the state-of-the-art methods. In the future, we would like to adopt attention mechanism to automatically select more discriminative body parts rather than simply slice the backbone feature $f_b$ alone the vertical direction.

\section*{Acknowledgment}
% National Key R\&D Program of China (No. 2018YFB1004600)
This work was supported by NSFC61573160 and NSFC61703171. This work was supported to Dr. Xiang Bai by the National Program for Support of Top-notch Young Professionals and the Program for HUST Academic Frontier Youth Team.

\section*{References}

\bibliography{longbibfile}

\begin{thebibliography}{10}
\expandafter\ifx\csname url\endcsname\relax
  \def\url#1{\texttt{#1}}\fi
\expandafter\ifx\csname urlprefix\endcsname\relax\def\urlprefix{URL }\fi
\expandafter\ifx\csname href\endcsname\relax
  \def\href#1#2{#2} \def\path#1{#1}\fi

\bibitem{zheng2017unlabeled}
Z.~Zheng, L.~Zheng, Y.~Yang, Unlabeled samples generated by gan improve the
  person re-identification baseline in vitro, in: Porc. of IEEE Intl. Conf. on
  Computer Vision, 2017, pp. 3774--3782.

\bibitem{zheng2015scalable}
L.~Zheng, L.~Shen, L.~Tian, S.~Wang, J.~Wang, Q.~Tian, Scalable person
  re-identification: A benchmark, in: Porc. of IEEE Intl. Conf. on Computer
  Vision, 2015, pp. 1116--1124.

\bibitem{li2014deepreid}
W.~Li, R.~Zhao, T.~Xiao, X.~Wang, Deepreid: Deep filter pairing neural network
  for person re-identification, in: Proc. of IEEE Intl. Conf. on Computer
  Vision and Pattern Recognition, 2014, pp. 152--159.

\bibitem{WuLDYBY19}
Y.~Wu, Y.~Lin, X.~Dong, Y.~Yan, W.~Bian, Y.~Yang, Progressive learning for
  person re-identification with one example, IEEE Trans. on Image Processing
  28~(6) (2019) 2872--2881.

\bibitem{FanZYY18}
H.~Fan, L.~Zheng, C.~Yan, Y.~Yang, Unsupervised person re-identification:
  Clustering and fine-tuning, ACM Transactions on Multimedia Computing,
  Communications, and Applications 14~(4) (2018) 83:1--83:18.

\bibitem{chen2017multi}
W.~Chen, X.~Chen, J.~Zhang, K.~Huang, A multi-task deep network for person
  re-identification., in: Proc. of the AAAI Conf. on Artificial Intelligence,
  2017, pp. 3988--3994.

\bibitem{li2017JLML}
S.~G. Wei~Li, Xiatian~Zhu, Person re-identification by deep joint learning of
  multi-loss classification, in: Proc. of Intl. Joint Conf. on Artificial
  Intelligence, 2017, pp. 2194--2200.

\bibitem{ustinova2017multi}
E.~Ustinova, Y.~Ganin, V.~Lempitsky, Multi-region bilinear convolutional neural
  networks for person re-identification, in: AVSS, 2017, pp. 1--6.

\bibitem{ahmed2015improved}
E.~Ahmed, M.~Jones, T.~K. Marks, An improved deep learning architecture for
  person re-identification, in: Proc. of IEEE Intl. Conf. on Computer Vision
  and Pattern Recognition, 2015, pp. 3908--3916.

\bibitem{BenGZJWM19}
X.~Ben, C.~Gong, P.~Zhang, X.~Jia, Q.~Wu, W.~Meng, Coupled patch alignment for
  matching cross-view gaits, IEEE Trans. on Image Processing 28~(6) (2019)
  3142--3157.

\bibitem{yao2017deep}
H.~Yao, S.~Zhang, R.~Hong, Y.~Zhang, C.~Xu, Q.~Tian, Deep representation
  learning with part loss for person re-identification, IEEE Trans. on Image
  Processing 28~(6) (2019) 2860--2871.

\bibitem{li2017learning}
D.~Li, X.~Chen, Z.~Zhang, K.~Huang, Learning deep context-aware features over
  body and latent parts for person re-identification, in: Proc. of IEEE Intl.
  Conf. on Computer Vision and Pattern Recognition, 2017, pp. 7398--7407.

\bibitem{wu2018what}
L.~Wu, Y.~Wang, X.~Li, J.~Gao, What-and-where to match: Deep spatially
  multiplicative integration networks for person re-identification, Pattern
  Recognition 76 (2018) 727--738.

\bibitem{zhao2017spindle}
H.~Zhao, M.~Tian, S.~Sun, J.~Shao, J.~Yan, S.~Yi, X.~Wang, X.~Tang, Spindle
  net: Person re-identification with human body region guided feature
  decomposition and fusion, in: Proc. of IEEE Intl. Conf. on Computer Vision
  and Pattern Recognition, 2017, pp. 907--915.

\bibitem{su2017pose}
C.~Su, J.~Li, S.~Zhang, J.~Xing, W.~Gao, Q.~Tian, Pose-driven deep
  convolutional model for person re-identification, in: Porc. of IEEE Intl.
  Conf. on Computer Vision, 2017, pp. 3980--3989.

\bibitem{varior2016siamese}
R.~R. Varior, B.~Shuai, J.~Lu, D.~Xu, G.~Wang, A siamese long short-term memory
  architecture for human re-identification, in: Proc. of European Conference on
  Computer Vision, 2016, pp. 135--153.

\bibitem{wu2016enhanced}
S.~Wu, Y.-C. Chen, X.~Li, A.-C. Wu, J.-J. You, W.-S. Zheng, An enhanced deep
  feature representation for person re-identification, in: Proc. of IEEE Winter
  Conf. on Applications of Computer Vision, 2016, pp. 1--8.

\bibitem{yan2018multi}
Y.~Yan, B.~Ni, J.~Liu, X.~Yang, Multi-level attention model for person
  re-identification, Pattern Recognition Letters.

\bibitem{zheng2016wild}
L.~Zheng, H.~Zhang, S.~Sun, M.~Chandraker, Y.~Yang, Q.~Tian, Person
  re-identification in the wild, in: Proc. of IEEE Intl. Conf. on Computer
  Vision and Pattern Recognition, 2017, pp. 3346--3355.

\bibitem{zheng2016person}
L.~Zheng, Y.~Yang, A.~G. Hauptmann, Person re-identification: Past, present and
  future, CoRR abs/1610.02984.
\newblock \href {http://arxiv.org/abs/1610.02984} {\path{arXiv:1610.02984}}.

\bibitem{wang2016joint}
F.~Wang, W.~Zuo, L.~Lin, D.~Zhang, L.~Zhang, Joint learning of single-image and
  cross-image representations for person re-identification, in: Proc. of IEEE
  Intl. Conf. on Computer Vision and Pattern Recognition, 2016, pp. 1288--1296.

\bibitem{liu2017end}
H.~Liu, J.~Feng, M.~Qi, J.~Jiang, S.~Yan, End-to-end comparative attention
  networks for person re-identification, IEEE Trans. on Image Processing 26~(7)
  (2017) 3492--3506.

\bibitem{hochreiter1997long}
S.~Hochreiter, J.~Schmidhuber, Long short-term memory, Neural computation 9~(8)
  (1997) 1735--1780.

\bibitem{cheng2016person}
D.~Cheng, Y.~Gong, S.~Zhou, J.~Wang, N.~Zheng, Person re-identification by
  multi-channel parts-based cnn with improved triplet loss function, in: Proc.
  of IEEE Intl. Conf. on Computer Vision and Pattern Recognition, 2016, pp.
  1335--1344.

\bibitem{wang2018equidistance}
J.~Wang, Z.~Wang, C.~Liang, C.~Gao, N.~Sang, Equidistance constrained metric
  learning for person re-identification, Pattern Recognition 74 (2018) 38--51.

\bibitem{zhao2017multiple}
C.~Zhao, X.~Wang, W.~K. Wong, W.~Zheng, J.~Yang, D.~Miao, Multiple metric
  learning based on bar-shape descriptor for person re-identification, Pattern
  Recognition 71 (2017) 218--234.

\bibitem{liu2018m3l}
X.~Liu, X.~Ma, J.~Wang, H.~Wang, M3l: Multi-modality mining for metric learning
  in person re-identification, Pattern Recognition 76 (2018) 650--661.

\bibitem{ZHAO201879}
C.~Zhao, X.~Wang, D.~Miao, H.~Wang, W.~Zheng, Y.~Xu, D.~Zhang, Maximal
  granularity structure and generalized multi-view discriminant analysis for
  person re-identification, Pattern Recognition 79 (2018) 79 -- 96.

\bibitem{CHENG2018}
D.~Cheng, Y.~Gong, X.~Chang, W.~Shi, A.~Hauptmann, N.~Zheng, Deep feature
  learning via structured graph laplacian embedding for person
  re-identification, Pattern Recognition.

\bibitem{ZHAO201890}
Z.~Zhao, B.~Zhao, F.~Su, Person re-identification via integrating patch-based
  metric learning and local salience learning, Pattern Recognition 75 (2018) 90
  -- 98, distance Metric Learning for Pattern Recognition.

\bibitem{KaranamGWRCR19}
S.~Karanam, M.~Gou, Z.~Wu, A.~Rates{-}Borras, O.~I. Camps, R.~J. Radke, A
  systematic evaluation and benchmark for person re-identification: Features,
  metrics, and datasets, IEEE Trans. Pattern Anal. Mach. Intell. 41~(3) (2019)
  523--536.

\bibitem{chen2016similarity}
D.~Chen, Z.~Yuan, B.~Chen, N.~Zheng, Similarity learning with spatial
  constraints for person re-identification, in: Proc. of IEEE Intl. Conf. on
  Computer Vision and Pattern Recognition, 2016, pp. 1268--1277.

\bibitem{fendri2018multi}
E.~Fendri, M.~Frikha, M.~Hammami, Multi-level semantic appearance
  representation for person re-identification system, Pattern Recognition
  Letters 115 (2018) 30--38.

\bibitem{pcb}
Y.~Sun, L.~Zheng, Y.~Yang, Q.~Tian, S.~Wang, Beyond part models: Person
  retrieval with refined part pooling (and {A} strong convolutional baseline),
  in: eccv, 2018, pp. 501--518.

\bibitem{jaderberg2015spatial}
M.~Jaderberg, K.~Simonyan, A.~Zisserman, et~al., Spatial transformer networks,
  in: Proc. of Advances in Neural Information Processing Systems, 2015, pp.
  2017--2025.

\bibitem{hermans2017defense}
A.~Hermans, L.~Beyer, B.~Leibe, In defense of the triplet loss for person
  re-identification, CoRR abs/1703.07737.
\newblock \href {http://arxiv.org/abs/1703.07737} {\path{arXiv:1703.07737}}.

\bibitem{geng2016deep}
H.~Chen, Y.~Wang, Y.~Shi, K.~Yan, M.~Geng, Y.~Tian, T.~Xiang, Deep transfer
  learning for person re-identification, in: Fourth {IEEE} Intl. Conf. on
  Multimedia Big Data, 2018, pp. 1--5.

\bibitem{qian2017multi}
X.~Qian, Y.~Fu, Y.-G. Jiang, T.~Xiang, X.~Xue, Multi-scale deep learning
  architectures for person re-identification, in: Porc. of IEEE Intl. Conf. on
  Computer Vision, 2017, pp. 5409--5418.

\bibitem{xu2015show}
K.~Xu, J.~Ba, R.~Kiros, K.~Cho, A.~Courville, R.~Salakhudinov, R.~Zemel,
  Y.~Bengio, Show, attend and tell: Neural image caption generation with visual
  attention, in: Proc. of Intl. Conf. on Machine Learning, 2015, pp.
  2048--2057.

\bibitem{sutskever2014sequence}
I.~Sutskever, O.~Vinyals, Q.~V. Le, Sequence to sequence learning with neural
  networks, in: Proc. of Advances in Neural Information Processing Systems,
  2014, pp. 3104--3112.

\bibitem{graves2013speech}
A.~Graves, A.~Mohamed, G.~E. Hinton, Speech recognition with deep recurrent
  neural networks, in: {IEEE} International Conference on Acoustics, Speech and
  Signal Processing, 2013, pp. 6645--6649.

\bibitem{shi2017end}
B.~Shi, X.~Bai, C.~Yao, An end-to-end trainable neural network for image-based
  sequence recognition and its application to scene text recognition, IEEE
  Trans. Pattern Anal. Mach. Intell. 39~(11) (2017) 2298--2304.

\bibitem{Wang2016unified}
J.~Wang, Y.~Yang, J.~Mao, Z.~Huang, C.~Huang, W.~Xu, Cnn-rnn: A unified
  framework for multi-label image classification, in: Proc. of IEEE Intl. Conf.
  on Computer Vision and Pattern Recognition, 2016, pp. 2285--2294.

\bibitem{bell2016inside}
S.~Bell, C.~Lawrence~Zitnick, K.~Bala, R.~Girshick, Inside-outside net:
  Detecting objects in context with skip pooling and recurrent neural networks,
  in: Proc. of IEEE Intl. Conf. on Computer Vision and Pattern Recognition,
  2016, pp. 2874--2883.

\bibitem{shi2016embedding}
H.~Shi, Y.~Yang, X.~Zhu, S.~Liao, Z.~Lei, W.~Zheng, S.~Z. Li, Embedding deep
  metric for person re-identification: A study against large variations, in:
  Proc. of European Conference on Computer Vision, 2016, pp. 732--748.

\bibitem{he2016deep}
K.~He, X.~Zhang, S.~Ren, J.~Sun, Deep residual learning for image recognition,
  in: Proc. of IEEE Intl. Conf. on Computer Vision and Pattern Recognition,
  2016, pp. 770--778.

\bibitem{ristani2016performance}
E.~Ristani, F.~Solera, R.~S. Zou, R.~Cucchiara, C.~Tomasi, Performance measures
  and a data set for multi-target, multi-camera tracking, in: ECCV Workshop on
  Benchmarking Multi-Target Tracking, 2016, pp. 17--35.

\bibitem{felzenszwalb2008discriminatively}
P.~Felzenszwalb, D.~McAllester, D.~Ramanan, A discriminatively trained,
  multiscale, deformable part model, in: Proc. of IEEE Intl. Conf. on Computer
  Vision and Pattern Recognition, 2008, pp. 1--8.

\bibitem{liao2015person}
S.~Liao, Y.~Hu, X.~Zhu, S.~Z. Li, Person re-identification by local maximal
  occurrence representation and metric learning, in: Proc. of IEEE Intl. Conf.
  on Computer Vision and Pattern Recognition, 2015, pp. 2197--2206.

\bibitem{lin2017improving}
Y.~Lin, L.~Zheng, Z.~Zheng, Y.~Wu, Y.~Yang, Improving person re-identification
  by attribute and identity learning, CoRR abs/1703.07220.
\newblock \href {http://arxiv.org/abs/1703.07220} {\path{arXiv:1703.07220}}.

\bibitem{zheng2017pedestrian}
Z.~Zheng, L.~Zheng, Y.~Yang, Pedestrian alignment network for large-scale
  person re-identification, IEEE Transactions on Circuits and Systems for Video
  Technology.

\bibitem{schumann2017person}
A.~Schumann, R.~Stiefelhagen, Person re-identification by deep learning
  attribute-complementary information, in: Proc. of IEEE Intl. Conf. on
  Computer Vision and Pattern Recognition Workshops, 2017, pp. 1435--1443.

\bibitem{sun2017svdnet}
Y.~Sun, L.~Zheng, W.~Deng, S.~Wang, Svdnet for pedestrian retrieval, in: Porc.
  of IEEE Intl. Conf. on Computer Vision, 2017, pp. 3820--3828.

\bibitem{chen2017person}
Y.~Chen, X.~Zhu, S.~Gong, Person re-identification by deep learning multi-scale
  representations, in: Porc. of IEEE Intl. Conf. on Computer Vision, 2017, pp.
  2590--2600.

\bibitem{zhou2017efficient}
J.~Zhou, P.~Yu, W.~Tang, Y.~Wu, Efficient online local metric adaptation via
  negative samples for person re-identification, in: Proc. of IEEE Intl. Conf.
  on Computer Vision and Pattern Recognition, 2017, pp. 2439--2447.

\bibitem{zhang2016learning}
L.~Zhang, T.~Xiang, S.~Gong, Learning a discriminative null space for person
  re-identification, in: Proc. of IEEE Intl. Conf. on Computer Vision and
  Pattern Recognition, 2016, pp. 1239--1248.

\bibitem{varior2016gated}
R.~R. Varior, M.~Haloi, G.~Wang, Gated siamese convolutional neural network
  architecture for human re-identification, in: Proc. of European Conference on
  Computer Vision, 2016, pp. 791--808.

\bibitem{chen2017personPAMI}
Y.-C. Chen, X.~Zhu, W.-S. Zheng, J.-H. Lai, Person re-identification by camera
  correlation aware feature augmentation, IEEE Trans. Pattern Anal. Mach.
  Intell. 40~(2) (2018) 392--408.

\bibitem{zhou2017point}
S.~Zhou, J.~Wang, J.~Wang, Y.~Gong, N.~Zheng, Point to set similarity based
  deep feature learning for person re-identification, in: Proc. of IEEE Intl.
  Conf. on Computer Vision and Pattern Recognition, 2017, pp. 5028--5037.

\bibitem{lin2017consistent}
J.~Lin, L.~Ren, J.~Lu, J.~Feng, J.~Zhou, Consistent-aware deep learning for
  person re-identification in a camera network, in: Proc. of IEEE Intl. Conf.
  on Computer Vision and Pattern Recognition, 2017, pp. 3396--3405.

\bibitem{zhao2017deeply}
L.~Zhao, X.~Li, Y.~Zhuang, J.~Wang, Deeply-learned part-aligned representations
  for person re-identification, in: Porc. of IEEE Intl. Conf. on Computer
  Vision, 2017, pp. 3239--3248.

\bibitem{bai2017scalable}
S.~Bai, X.~Bai, Q.~Tian, Scalable person re-identification on supervised
  smoothed manifold, in: Proc. of IEEE Intl. Conf. on Computer Vision and
  Pattern Recognition, 2017, pp. 3356--3365.

\bibitem{matsukawa2016hierarchical}
T.~Matsukawa, T.~Okabe, E.~Suzuki, Y.~Sato, Hierarchical gaussian descriptor
  for person re-identification, in: Proc. of IEEE Intl. Conf. on Computer
  Vision and Pattern Recognition, 2016, pp. 1363--1372.

\bibitem{cheng2017DictRW}
D.~Cheng, X.~Chang, L.~Liu, A.~G. Hauptmann, Y.~Gong, N.~Zheng, Discriminative
  dictionary learning with ranking metric embedded for person
  re-identification, in: Proc. of Intl. Joint Conf. on Artificial Intelligence,
  2017, pp. 964--970.

\bibitem{chen2017beyond}
W.~Chen, X.~Chen, J.~Zhang, K.~Huang, Beyond triplet loss: A deep quadruplet
  network for person re-identification, in: Proc. of IEEE Intl. Conf. on
  Computer Vision and Pattern Recognition, 2017, pp. 1320--1329.

\bibitem{zhong2017re}
Z.~Zhong, L.~Zheng, D.~Cao, S.~Li, Re-ranking person re-identification with
  k-reciprocal encoding, in: Proc. of IEEE Intl. Conf. on Computer Vision and
  Pattern Recognition, 2017, pp. 3652--3661.

\end{thebibliography}

\end{document}